# COLOR TEAMS FOR MACHINE LEARNING DEVELOPMENT

Josh Kalin[1,2], David Noever[2], Matthew Ciolino[2]

[1]Department of Computer Science and Software Engineering, Auburn University, Auburn, AL, USA
jzk0098@auburn.edu

[2]PeopleTec, Inc, Huntsville, AL, USA
David.Noever@peopletec.com, Matt.Ciolino@peopletec.com

## ABSTRACT

Machine learning and software development share processes and methodologies for reliably delivering products to customers. This work proposes the use of a new teaming construct for forming machine learning teams for better combatting adversarial attackers. In cybersecurity, infrastructure uses these teams to protect their systems by using system builders and programmers to also offer more robustness to their platforms. Color teams provide clear responsibility to the individuals on each team for which part of the baseline (Yellow), attack (Red), and defense (Blue) breakout of the pipeline. Combining colors leads to additional knowledge shared across the team and more robust models built during development. The responsibilities of the new teams Orange, Green, and Purple will be outlined during this paper along with an overview of the necessary resources for these teams to be successful.

## KEYWORDS

*Neural Networks, Machine Learning, Image Classification, Adversarial Attacks*

## INTRODUCTION

This note seeks to address the construction of a machine learning development team. Traditionally, machine learning teams focus on investigating machine learning applications in practical spaces, building promising architectures, and finally deploying the best models to production. Teams can take on all of these aspects or just focus on one at a time. Machine learning adoption in the industry affects every facet of software development from code generation to testing [1]. The production-focused models are then evaluated for performance in normal and adversarial settings. In this work, we propose formally assigning responsibilities to individual parts of the team to focus on their core strength.

## COLOR TEAMS FOR MACHINE LEARNING DEVELOPMENT

Cremen introduced the InfoSec Color Wheel for structuring teams in the Cyber-Physical and Software systems [2]. The goal of this work is to apply the same concepts with machine learning developers in mind and build more robust models from the beginning of the development process. Focusing on including the attackers in the development loop allows the entire team to understand the holes and vulnerabilities in the included system. Models are so large now that it is nearly impossible to supply explanations for each sample without massive computer resources or time delay in the answers created [3]. Due to the complexity of explaining machine learning models, it is necessary to form teams that can baseline, attack, and defend ML models.

## DESIGNING THE TEAMS: FAMILIAR CONCEPTS

Cybersecurity commonly uses the Red, Blue, and Yellow teams to break out clear lines of responsibility during the hardening of cyber-physical and software systems [4]. Each team has a distinct role to play in baselining, attacking, or defending the system as shown in Figure 1. In machine learning development, these concepts generally translate to model selection, parameter reduction, and pruning or data augmentation to protect those models [5]. Each model development pipeline though needs a specific plan for combatting known and predictable vulnerabilities that an attacker could exploit [6].

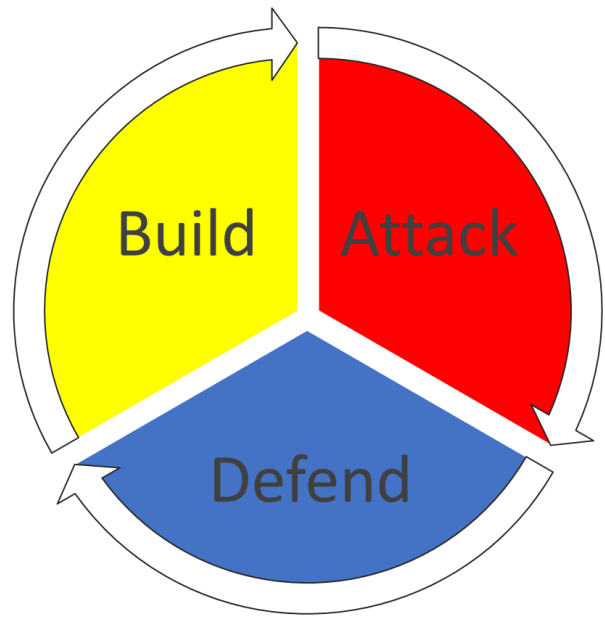

*Figure 1: Build, Attack, Defend is a common model in cyber security and is emerging as a method to protect machine learning models in development [5].*

### Yellow Team: Build Phase
Yellow Team or the Development Team is responsible for building the product that is attacked and defended. In the Cyber Security domain, this is building the cyber-physical or software system [7]. In Machine Learning, these are the model builders and creators that train, build, and deploy the model to the appropriate systems [8]. The Yellow Team is solely responsible for the creation and deployment of the model.

### Red Team: Attack Phase
The Red Team serves as system attackers. This team understands known vulnerabilities and exploitative methods including how they may be used against your system. An expectation for the red team is its ability to attack a system and provide a list of exploits that work on the current deployment [9]. In some cases, the Red Team can also supply suggestions for how to circumvent these issues. For machine learning, these attacks can range from the deployment methodology to known issues with the architecture or datasets [10].

### Blue Team: Defend Phase
Red Team attacks and Blue Team defends. The Blue Team understands the known exploits and vulnerabilities for the systems deployed but introduces solutions for the development team to implement [11]. In the machine learning context, this function would pertain to the use of machine learning libraries and the selection of architectures and datasets.

## Responsibilities and Roles

Each team in the classic configuration has defined and clear responsibilities in the development of their systems. In recent years, it has become apparent that a lack of communication and understanding of vulnerabilities in common development platforms has led to a need for blending these teams into mixtures of colors [12]. The next section will map these additional team constructs into the machine learning development lifecycle.

## ADDING NEW TEAM CONFIGURATIONS

There are three additional team constructs typically applied to the cybersecurity field: Orange, Purple, and Green. The focus of each team is to include attacker knowledge directly into the team to ensure that each step includes some protections from the beginning. The focus of this section is to describe the mapping of a machine learning development process to these new team constructs.

### Orange Team

As illustrated in Figure 2, the Orange Team is focused on bringing organizational changes to a team around building more robust architectures and datasets. In a classic configuration, the teams constructing datasets or developing the models may not worry about attacks on those deployments. The Orange team is establishing new processes and educating developers on best practices to prevent adversarial attackers from launching successful attacks. In smaller organizations, this role could be taken on by a single person with development and adversarial machine learning knowledge.

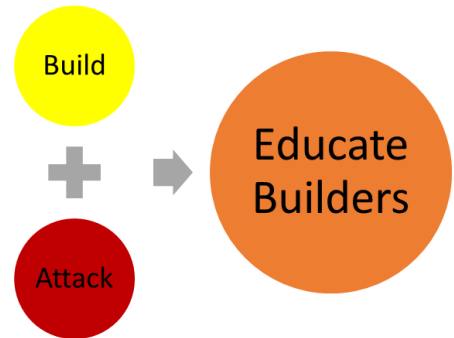

*Figure 2: Orange Team educates builders and creates robust ML design patterns for development*

### Purple Team

As shown in Figure 3, the Purple Team integrates defensive tactics based on the adversarial results and continues to improve processes established by the Orange Team. Each attack brings a set of solutions and issues that might prevent that particular attack from being successful. Purple Team works with both Blue and Red teams to develop strategies for understanding the adversarial surface of a model and how to protect both the model and datasets from those attackers. If Orange Team is responsible for maintaining processes and educating the team, then Purple Team is responsible for keeping models robust from attacks.

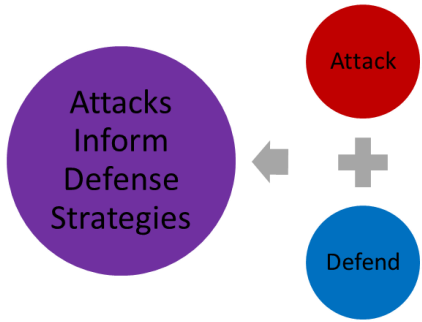

*Figure 3: Purple Team creates defense strategies from attacks launched on target models*

### Green Team

As shown in Figure 4, the Green Team will integrate best practices into the development of the models and dataset. As developers continue to understand vulnerabilities, they begin to integrate, use, and apply these new techniques into their new processes. The models will be evaluated against a statistical model for assessing adversarial risk like the modified Drake equation [13] to understand the risk factors in the current model architectures. Finally, the Green Team will continue to improve the model over the development cycle from concept to production deployment. The Green Team will use strategies and patterns designed by both Purple and Orange Team members.

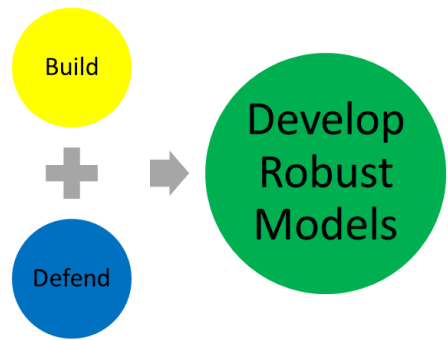

*Figure 4: Green Team builds secure models by utilizing robust design patterns with informed defense strategies*

## ACCOUNTING FOR THE PROCESSES

Traditional software teams rely on multiple decades of lessons learned to produce efficient code on flexible timelines. With machine learning and data science teams, there is a longer learning curve if solving problems with techniques borrowed from the development and business side [14]. As shown in Figure 5, this section covers the six steps in the overall process: Strategy, Design, Development, Testing, Deployment, and Maintenance.

### Strategy

Each machine learning development team will need to create, maintain, and update a strategy for building and maintaining effective models in their production environments. The strategy section will incorporate the teaming concepts to add value at each development step. For instance, in traditional Development Operations, developers would be concerned with choosing the correct data structures, libraries, and methodologies without considering the security ramifications of each of those choices. Development Security Operations (DevSecOps) is a new process by which the developers actively design techniques that are more secure from the input [15-16]. In the same vein, we introduce Machine Learning Security Operations or MLSecOps as another pathway for protecting models [17]. By using these teaming constructs, a machine learning team can build adversarial protections into their model development from the beginning of the strategy phase.

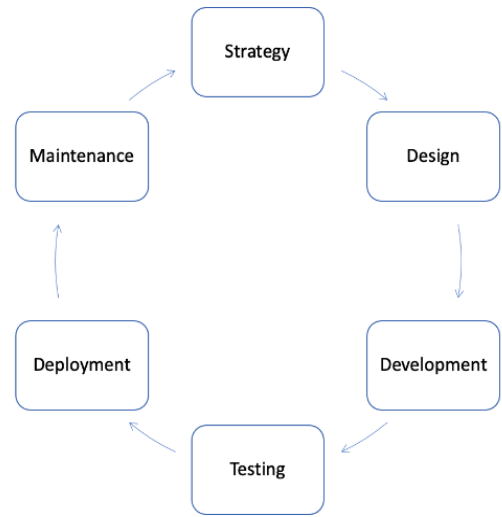

*Figure 5: The development of machine learning models is a continuous loop of building new and evolving strategies to build, baseline, and defend models in production environments*

### Design

The design phase is reserved for understanding the proper structures, methodologies, inputs, outputs, and general operations of the requirements being built with software. In a machine learning team, the data inputs and model architectures selected can have a large impact on the adversarial risk assigned to that particular development cycle. By using the Green Team to inform proper datasets and architectures, the Green and Yellow Team together can design more robust models from the beginning. Cross-functional inclusion of team members (team members can belong to more than one group) will lead to additional lessons learned between the steps in these processes. Design, in a machine learning team, should be focused on solving the problem with a robust, secure solution that has the best chance of maintaining a longer lifespan when deployed to Production [18].

### Development

In traditional software teams, development can be straightforward and deterministic for tasks like building frontend or backend code for websites. While there are challenges with the implementation of tools or algorithms, estimating the number of story points can become deterministic with Agile teams [19-20]. In machine learning development, the methods for estimating stories do require some updating. Singla et.al. analyzed several machine learning and non-machine learning projects to understand the difference in estimating and planning for ML-based projects [21]. In their analysis, they discovered that ML teams had more challenges with completing stories but the success stories included descriptive titles, clear labeling for the ML domain used, and measurable "Done" criteria in Agile methodology. Overall, this demonstrated that the development process for ML projects has commonality with normal development projects with stricter adherence to Agile tenants such as clear story descriptions and "Done" criteria.

*Testing*

Test suites are commonly structured around the unit- and integration-testing of software functionality. DevSecOps also adds testing for best security practices in coding and deployment [22]. Similarly, for a machine learning project, a common suite of testing tools is needed for evaluating machine learning models for susceptibility to incoming attacks. Each incoming attack can be tailored to an incoming architecture and, in the same vein, a Purple Team can devise a set of standard attacks for a team to develop or defend against. The Green Team then develops strategies and solutions for protecting models from these incoming attacks by using the Purple Team's recommended attacks. Each team has responsibility for mitigating attacks while still meeting performance targets. If performance cannot be met due to fixes implemented, then those attacks must be carried forward as the risk for review boards in a production team.

*Deployment*

The Green Team focuses on the task of defending a model while also building to meet performance targets. After testing, the Green Team adjudicates the outstanding risks and prepares for a release candidate for the machine learning model. The model is deployed to production with proper evaluation of the available exploits and each of those risks is properly included within the model documentation for the release. Hotfixes for a release are part of the deployment phase as certain aspects of the model environment or the evaluation may lead to additional mitigations like needing a version of a package or a way of processing data to avoid exploits from adversarial actors.

*Maintenance*

Every machine learning project will have the burden of maintaining the models in the face of new and changing data. To maintain a machine learning model, the deployment platform must monitor the incoming data and detect data drift in the original model. If a population difference is detected, the model will need to be retrained and redeployed. This maintenance event needs to have periodic evaluation and downtime built into the system. Like monitoring the dataset drift and its distribution, an additional step is added to detect possible attacks. A security patch in the machine learning world may look a lot like a new model – except the goal of the updated model is to mitigate the effectiveness of the newly detected attack. A running list of security exploits should be maintained and detected so attackers can be excluded or banned from using the deployed resource.

## ALLOCATING THE RESOURCES

Machine learning projects typically have four major hurdles during development: available computer resources, available Graphical Processing Units (GPUs), appropriate time allotted, and personnel shortages. Computer resources, including GPU, are used to process data, train models, and evaluate performance. In this section, each of the major components will build on the next. Note: there are many configurations of CPU and GPU assets in a team (Figure 6). This work represents one such allocation of those resources.

*Computer Resources*

Each development cycle will include an allotment of computing resources for performing the task. To combat adversarial actors and their techniques, the Green Team will need to employ state-of-the-art (SoTA) processes to improve the model. Adding additional steps like parameter search and neural architecture search for improving model performance can add burden to available compute. The team will need to evaluate the risk and reward of making the model more robust to attacks.

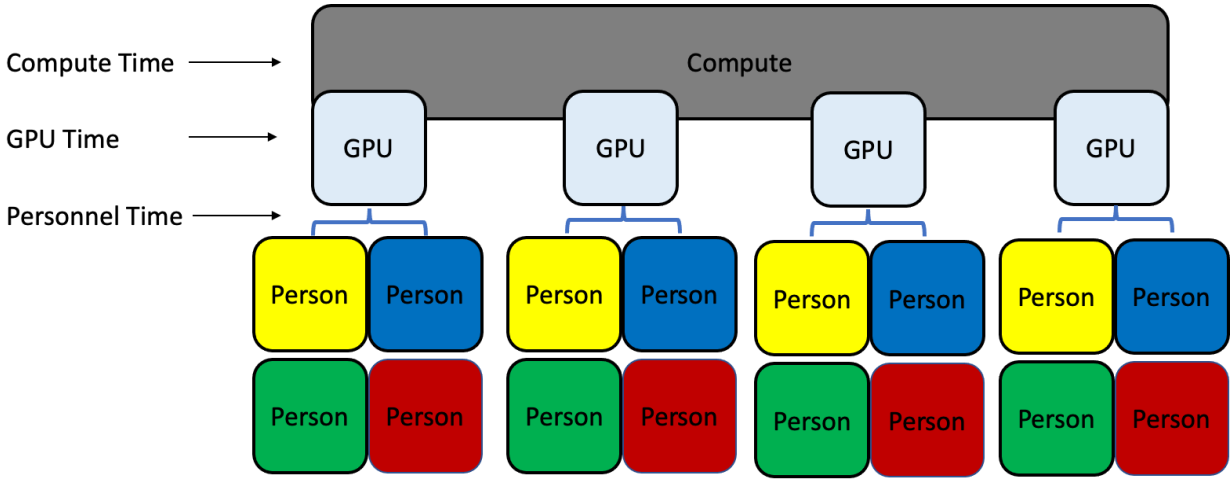

*Figure 6: Summary of key resource allocation steps to move a project from conception, training, testing and deployment. The personnel colors reflect the color team task and possible alignment with project plans.*

*GPUs*
In small to medium organizations, GPUs are a resource for training deep networks and transformers. As transformers begin to take over many tasks in the field of machine learning, the high VRAM GPUs are necessary for training and inference. Due to the prohibitive cost of cloud computing for higher-end GPUs, it's necessary for organizations, who have security concerns, to create and maintain internal on-premise GPU clusters. As a result, rationing GPU use to the different targets will guide how to best apply them and what jobs get higher priorities.

*Personnel*
The compute, GPUs, and time are all reliant on available time from personnel to utilize those resources in each of the color teams. The personnel side of the machine learning development team is a challenging task when there are multiple teams and competing priorities for each project. In using this structure for building color teams with tailored tasking for protecting models, it is important to cross-train the personnel on each color team's responsibilities to understand each of the different development requirements.

*Time*
Model development timelines include acquisition of the data, preprocessing of the data, building the model, and optimizing for a given task. Adding a time component of protecting these models will add complexity which can affect the schedule and available resources. To compound the issue, it can take considerably more resources to add additional robustness to a model with current SoTA techniques for protecting machine learning models. Three key time-related parameters can be monitored to support effective resource usage:

- *Compute Time*
    - Compute time represents the entire machine use time including memory, CPU time, GPU time, and data storage
- *GPU Time*
    - A specific mechanism for tracking the GPU usage as defending models can involve parameter and network-tuning to avoid certain types of attacks using, for instance, adversarial training methods on augmented datasets
- *Personnel Time*
    - Tracking personnel available to build, attack, and defend models is needed to fix high-priority risks in production systems

*SUMMARY AND FUTURE WORK*

The theoretical rate-limiting step is the Yellow Team which builds the time-consuming initial code. Every iteration of the code is going to improve its defenses against adversarial attacks. The Yellow Team is charged with building the initial set of software that can meet a mix of demands between protecting the model from attackers and meeting the customer requirements for performance. These tradeoffs can be difficult to manage as performance and defense are typically at odds with each other. As the performance of a model increases, it can be more susceptible to incoming attacks. Likewise, as the defendability of the model increases, it can be difficult to meet performance targets like speed, accuracy, and explainability.

Each step in this process is designed to create a team that can manage the demands of creating production-ready machine learning models with security in mind. As a team develops each new model, they will account for new and emerging threats in their model space. Drawing on the cybersecurity analogy, examples like a formal framework (MITRE ATT&CK [23]) should be developed for each color team. For instance, the Red Team catalogs successful attacks based on the vulnerability while the Green Team similarly logs the appropriate remediation or response. In this way, the rotation of personnel skills and time management may benefit future collaborative efforts between colors and project goals.

The future work of this research should plan and model each color team's steps to find the rate-limiting areas and where optimizations can be found. The color teams proposed in this work can be implemented as a system or in pieces that fit the needs of each project. This is a framework for defending deployed models from adversarial attacks in the wild.

## ACKNOWLEDGMENTS

The authors would like to thank PeopleTec's Technical Fellows program for its encouragement and project assistance. The views and conclusions contained in this paper are those of the authors and should not be interpreted as representing any funding agencies.

## REFERENCES

[1] Angelopoulos, Angelos, et al. "Tackling faults in the industry 4.0 era—a survey of machine-learning solutions and key aspects." Sensors 20.1 (2020): 109.
[2] Hackernoon, Accessed 03 January 2021, https://hackernoon.com/introducing-the-infosec-colourwheel-blending-developers-with-red-and-blue-security-teams-6437c1a07700
[3] Tjoa, Erico, and Cuntai Guan. "A survey on explainable artificial intelligence (XAI): Toward medical XAI." IEEE Transactions on Neural Networks and Learning Systems (2020).
[4] Seker, Ensar, and Hasan Huseyin Ozbenli. "The concept of cyber defence exercises (CDX): Planning, execution, evaluation." *2018 International Conference on Cyber Security and Protection of Digital Services (Cyber Security)*. IEEE, 2018.
[5] Vartak, Manasi, and Samuel Madden. "Modeldb: Opportunities and challenges in managing machine learning models." *IEEE Data Eng. Bull.* 41.4 (2018): 16-25.
[6] Insua, David Rios, et al. "Adversarial machine learning: Perspectives from adversarial risk analysis." *arXiv preprint arXiv:2003.03546* (2020).
[7] Kalin, Josh, David Noever, and Gerry Dozier. "Systematic attack surface reduction for deployed sentiment analysis models." *arXiv preprint arXiv:2006.11130* (2020).
[8] Russo, Lorenzo, Francesco Binaschi, and Alessio De Angelis. "Cybersecurity exercises: Wargaming and red teaming." *Next Generation CERTs* 54 (2019): 44.
[9] Baier, Lucas, Fabian Jöhren, and Stefan Seebacher. "Challenges in the deployment and operation of machine learning in practice." (2019).
[10] Kont, Markus, et al. "Frankenstack: Toward real-time red team feedback." *MILCOM 2017-2017 IEEE military communications conference (milcom)*. IEEE, 2017.

[11] Anderson, Hyrum. "The Practical Divide between Adversarial {ML} Research and Security Practice: A Red Team Perspective." (2021).
[12] Diogenes, Yuri, and Erdal Ozkaya. *Cybersecurity??? Attack and Defense Strategies: Infrastructure security with Red Team and Blue Team tactics*. Packt Publishing Ltd, 2018.
[13] Spring, Jonathan M., et al. "On managing vulnerabilities in AI/ML systems." New Security Paradigms Workshop 2020. 2020.
[14] Kalin, Josh, David Noever, and Matthew Ciolino. "A Modified Drake Equation for Assessing Adversarial Risk to Machine Learning Models." arXiv preprint arXiv:2103.02718 (2021).
[15] Khomh, Foutse, et al. "Software engineering for machine-learning applications: The road ahead." *IEEE Software* 35.5 (2018): 81-84.
[16] Myrbakken, Håvard, and Ricardo Colomo-Palacios. "DevSecOps: a multivocal literature review." *International Conference on Software Process Improvement and Capability Determination*. Springer, Cham, 2017.
[17] Mirsky, Yisroel, et al. "The Threat of Offensive AI to Organizations." *arXiv preprint arXiv:2106.15764* (2021).
[18] Washizaki, Hironori, et al. "Studying software engineering patterns for designing machine learning systems." *2019 10th International Workshop on Empirical Software Engineering in Practice (IWESEP)*. IEEE, 2019.
[19] Amershi, Saleema, et al. "Software engineering for machine learning: A case study." *2019 IEEE/ACM 41st International Conference on Software Engineering: Software Engineering in Practice (ICSE-SEIP)*. IEEE, 2019.
[20] Ungan, Erdir, Numan Cizmeli, and Onur Demirörs. "Comparison of functional size-based estimation and story points, based on effort estimation effectiveness in SCRUM projects." *2014 40th EUROMICRO Conference on Software Engineering and Advanced Applications*. IEEE, 2014.
[21] Singla, Kushal, et al. "Story and Task Issue Analysis for Agile Machine Learning Projects." 2020 IEEE-HYDCON. IEEE, 2020.
[22] Hsu, Tony Hsiang-Chih. *Practical security automation and testing: tools and techniques for automated security scanning and testing in DevSecOps*. Packt Publishing Ltd, 2019.
[23] Strom, B. E., Applebaum, A., Miller, D. P., Nickels, K. C., Pennington, A. G., & Thomas, C. B. (2018). Mitre att&ck: Design and philosophy. Technical report.

## AUTHORS

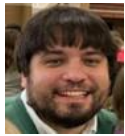
**Josh Kalin** is a physicist and data scientist focused on the intersections of robotics, data science, and machine learning. Josh holds undergraduate and graduate degrees in Physics, Mechanical Engineering, and Computer Science.

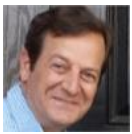
**David Noever** has research experience with NASA and the Department of Defense in machine learning and data mining. He received his Ph.D. from Oxford University, as a Rhodes Scholar, in theoretical physics.

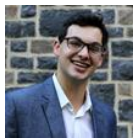
**Matt Ciolino** has experience in deep learning and computer vision. He received his Bachelor's from Lehigh University in Mechanical Engineering.